# Model-Based Hierarchical Clustering

Shivakumar Vaithyanathan and Byron Dom
IBM Almaden Research Center
650 Harry Rd.
San Jose, CA 95120-6099

## Abstract

We present an approach to model-based hierarchical clustering by formulating an objective function based on a Bayesian analysis. This model organizes the data into a cluster hierarchy while specifying a complex feature-set partitioning that is a key component of our model. Features can have either a unique distribution in every cluster or a common distribution over some (or even all) of the clusters. The cluster subsets over which these features have such a common distribution correspond to the nodes (clusters) of the tree representing the hierarchy. We apply this general model to the problem of document clustering for which we use a multinomial likelihood function and Dirichlet priors. Our algorithm consists of a two-stage process wherein we first perform a flat clustering followed by a modified hierarchical agglomerative merging process that includes determining the features that will have common distributions over the merged clusters. The regularization induced by using the marginal likelihood automatically determines the optimal model structure including number of clusters, the depth of the tree and the subset of features to be modeled as having a common distribution at each node. We present experimental results on both synthetic data and a real document collection.

## 1 Introduction

Recent years have seen significant interest in model-based clustering, in which it is assumed that the data is generated by a mixture of underlying probability distributions where each of the components can be interpreted as a cluster. A complete description of such model-based clustering can be found in [1]. Extensions to these generative models incorporating hierarchical agglomerative algorithms have also been studied[6]. These algorithms operate by merging clusters such that the resulting likelihood is maximized. Efficient algorithms for model-based hierarchical clustering of special cases of Gaussian generative models are described in Fraley [6]. In addition a model-based HAC algorithm based on a multinomial mixture model has been developed[9]. In the rest of the paper our references to HAC will be to the version of HAC used in a likelihood setting as described above. In particular we will be concentrating on multinomial mixture models.

Other hierarchical clustering algorithms in the literature include [8], which describes a scheme to characterize text collections hierarchically based on a deterministic annealing algorithm. In this model, besides the latent variables used for clustering the documents at the base of the hierarchy, additional latent variables are used to define intermediate nodes. These additional latent variables (called abstraction nodes) model the conditional probabilities of the words. Regularization is achieved by maximizing the likelihood on a separate validation data set. As will be seen, our model differs considerably from the work described in [8]. First, only a part of the feature set is modeled at the intermediate nodes in the hierarchy (the choice of these features is a part of the model selection). This can dramatically reduce the number of adjustable parameters in the model, especially in the context of very high-dimensional data. Second, regularization is achieved by integrating over the parameter values in a Bayesian fashion. This allows the model to utilize the entire data-set in the model building (i.e. no held-out validation set is required). Further, the two-stage algorithm described in this paper uses only the sufficient statistics computed in the first stage during the merging process and therefore can be computationally less expensive.

The probabilistic model developed in this paper orga-nizes the data elements into a natural hierarchy. The nodes at the different levels of the hierarchy are determined by modeling *features* that are common across the children of a particular node. These common features are modeled as having the same distribution across the clusters, based on the idea that some features, while having such a common distribution across some clusters, can be useful in discriminating among other clusters. We call these common features "noise" features at that node and the discriminatory features "useful". Such an analysis leads to a natural hierarchical model and lends itself nicely to some real problems like automatic taxonomy generation from a large collection of documents. A more formal definition of these *noise* and *useful* features is provided in subsequent sections. In passing we note that document clustering is a rich area with several approaches such as [12, 5]

The next section describes this hierarchical model and also derives the objective function, based on marginal likelihood, that describes such a model. In Section 3 we describe some approximate schemes to solve this objective function. Section 4 discusses the experimental set-up, results and evaluation. Finally Section 5 describes some future work in this area.

## 2 Models for Unsupervised Learning

In previous work[14, 13] we addressed the problem of *flat* (non-hierarchical) *partitional* (each data element belongs to one and only one cluster) clustering. Here we first describe that model in general terms and then extend it to the case of hierarchical, *partitional* clustering.

### 2.1 Model for Flat Partitional Clustering

A key aspect of this previous model is the partitioning of the features into two sets (with associated subspaces) - "noise" ($N$) and "useful" ($U$), with the following definitions.

**noise** features have the same statistical distribution across all clusters. Thus they would be useless as discriminators between clusters.

**useful** features, on the other hand, have a different distribution for each cluster and are therefore potentially useful for discrimination.

Our previous model has only these two feature types, precluding the possibility of features having the same distribution in some, but not all, of the clusters. Having said this, we must add that it is only true in a certain sense. In principle two features could have identical observed statistics in two clusters, resulting in identical parameter estimates. Despite these identical values, however, the two are treated as distinct and thus each have their own Bayesian prior distribution and will be "charged" as two distributions for regularization purposes. This issue will be discussed at length in the sequel.

This flat-clustering model (for the probability of the data $D$ conditioned on the *model structure* $\Omega$) can be written (in general terms) as:

$$P(D|\Omega) = P(D^N|\Omega) \prod_{k \in \Psi} P_k(D_k^U|\Omega), \qquad (1)$$

where

$D$ is the complete data set in the full feature representation.

$\Omega$ represents the model *structure*, which is specified by: (1) the partition of the complete feature set into $N$ and $U$, (2) the number of clusters and, (3) the assignment of the individual data items $\{d_i\}$ to clusters.

$D^N$ represents $D$ projected onto the *noise* subspace $N$.

$\Psi$ is the set of clusters, indexed by $k$.

$D_k^U$ represents the data elements associated with cluster $k$, $D_k$, projected onto the *useful* subspace $U$.

$P_k$ is the specific probability model associated with cluster $k$.

In [14, 13] we applied the general model form of (1) to the problem of document clustering, using *multinomial* distributions and the Bayesian[1] marginal likelihood (computed using Dirichlet distributions as priors) was the resulting objective function $P(D|\Omega)$. In this paper we use the same models for document clustering, but we defer that discussion until after we have developed the general form (corresponding to (1)) for the hierarchical model.

### 2.2 Model for Hierarchical Partitional Clustering

We now extend the flat-clustering model to the case of a cluster hierarchy (represented by a tree), in which all nodes correspond to clusters and all nodes except the leaves are further decomposed into other clusters and so on recursively until the leaf clusters are reached. A simple example of such a cluster hierarchy is diagrammed in Figure 1. We note here that a model similar to ours has been discussed in [7].

---

[1] See [3] for an in-depth treatment of the Bayesian statistical paradigm.

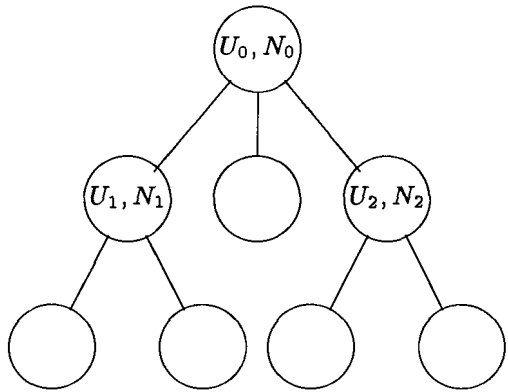

Figure 1: Diagram of a simple taxonomy showing feature-set partitions

To do this extension we will need to first extend the notation used in (1). The various symbols for sets, functions, parameters and so on will be indexed by the hierarchy node to which they apply. Thus:

- $U, N \rightarrow U_k, N_k$, which are the *useful* and *noise* features associated with hierarchy node $k$. Some further definition is required for these. This is discussed below. The index value $k = 0$ corresponds to the root node which contains all of $D$ (a.k.a. $D_0$).

- $\Psi \rightarrow \Psi_k$, the *children* of node $k$.

Our hierarchical model also extends the *noise/useful* concept to the more general case alluded to above. That is, the case where certain features can have a common distribution over some proper subset of all the clusters. In our model not all cluster subsets are candidates for such common distributions. Such a common feature distribution is associated with a node in the hierarchy and applies only to that node and all its descendants (i.e. the *entire sub-tree* beneath that node). The associated features are represented by $N_k$. Note that if node $j$ is an ancestor of node $k$, then $N_j \subseteq N_k$. Let $p(k)$ be the parent of $k$; we define:

$$\tilde{N}_k \equiv N_k - N_{p(k)}. \qquad (2)$$

The complement of $N_k$ is denoted by $U_k$, the "useful" features at node $k$.

$$U_k + \tilde{N}_k = N_{p(k)}. \qquad (3)$$

To construct the hierarchical model corresponding to (1) we substitute the extended notation yielding:

$$P(D|\Omega) = P(D_0^{N_0}|\Omega) \prod_{k \in \Psi_0} P_k(D_k^{U_0}|\Omega), \qquad (4)$$

Then we recursively expand the $P_k(D_k^{U_0}|\Omega)$ terms:

$$P_k(D_k^{U_0}|\Omega) = P(D_k^{\tilde{N}_k}|\Omega) \prod_{\ell \in \Psi_k} P(D_\ell^{U_k}|\Omega). \qquad (5)$$

For any non-leaf node $j$ in the hierarchy, we expand similarly:

$$P_k(D_k^{U_{p(k)}}|\Omega) = P(D_k^{\tilde{N}_k}|\Omega) \prod_{\ell \in \Psi_k} P(D_\ell^{U_k}|\Omega), \qquad (6)$$

The completely expanded form for $P(D|\Omega)$ for the entire hierarchy, obtained by recursively applying (6) until leaves are reached, will be quite complicated for complex cluster hierarchies and all the more so once specific forms are substituted for the $P_k$'s.

### 2.3 Marginal Likelihood of the Hierarchy

The final forms we use for $P(D|\Omega)$ and its various components (e.g. $P(D_k^{U_{p(k)}}|\Omega)$) will be *Bayesian marginal likelihoods*. That is, they will be based on an underlying distribution, for example the *multinomial*, which we use to address the application of document clustering, and an associated *prior* distribution (Dirichlet in the case of multinomials).

Most distributions are characterized by a set of real-valued parameters, which we represent by $\xi$. In the Bayesian paradigm these are treated as random variables with their own distribution, referred to as a "prior", which we represent by $\pi(\xi|\Omega)$. A single component distribution is specified by values of $\Omega$ and $\xi$. The *marginal likelihood* is formed by integrating the joint distribution $P(D, \xi|\Omega) = P(D|\xi, \Omega)\pi(\xi|\Omega)$ over the domain of $\xi$:

$$P(D|\Omega) = \int P(D|\xi, \Omega)\pi(\xi|\Omega)d\xi$$

Thus, in our model, the various terms in (1) through (6) will be expanded this way following the recursive hierarchical partitioning of the feature set described in Subsection 2.2. This expansion is complete at the leaves of the hierarchy.

### 2.4 A Multinomial-Based Model for Hierarchical Document Clustering

To use this general approach to construct a model for a hierarchical partition of a set of documents, we start by treating a document as a "bag of words" (as is common practice) thus ignoring any information about the sequence of terms other than how many times each term appears. These term counts are the features we use. This treatment is often referred to in information retrieval parlance as the "vector-space" model. We further assume that the probabilities of occurrence of all terms are independent of how many times the terms themselves or any other terms occur in the same or any other document. Despite the obvious incorrectness of this assumption, for text, it is also commonly made and has produced quite satisfactory results. The implication of these assumptions is that the statistical distribution of these terms counts is the *multinomial*.

### 2.4.1 Model for a Single Cluster

The probability of observing a single document $d$ composed of words from an $M$-term lexicon with term counts $\{t_j | j = 1, 2, \ldots, M\}$ is:

$$P(d|\xi) = \binom{n}{\{t_j\}} \prod_{j=1}^{M} \theta_j^{t_j}. \quad (7)$$

where $\binom{n}{\{...\}}$ is a multinomial coefficient, $n = \sum_{j=1}^{M} t_j$ and $\theta_j$ is the probability of a word being the $j^{th}$ term of the lexicon. We identify the set $\{\theta_j\}$ with $\xi$ of Subsection 2.3. The probability of a collection $D$ of $\nu$ independent documents $\{d_i | i = 1, 2, \ldots, \nu\}$ from the same population (e.g. a single cluster) is described by a product of terms like (7):

$$p(D|\xi) = \prod_{i=1}^{\nu} \binom{n_i}{\{t_{i,j}\}} \prod_{j=1}^{M} \theta_j^{t_{i,j}} \quad (8)$$

where the notation of (7) has been augmented in the obvious way with the document index $i$.

To obtain the marginal likelihood for this collection of documents we integrate the product of (8) and the prior over the domain of $\{\theta_j\}$, which is the simplex in $\mathbb{R}^M$ defined by $\forall_j \theta_j \in [0, 1]$ and the normalization constraint $\sum_{j=1}^{M} \theta_j = 1$. As stated above our choice of prior is the *Dirichlet* distribution[3]:

$$\pi(\theta) = \frac{\Gamma(\alpha_0)}{\prod_{i=1}^{M} \Gamma(\alpha_i)} \prod_{j=1}^{M} \theta_j^{(\alpha_j - 1)}, \quad (9)$$

where $\alpha_0 = \sum_{j=1}^{M} \alpha_i$ and the $\{\alpha_i\}$ are referred to as *hyperparameters*. This integration results in the following expression, which is a product of terms that are instances of what is known as the *Multinomial-Dirichlet* distribution[2]. The marginal likelihood can be written as below:

$$p(D) = \frac{\Gamma(\alpha_0)}{\Gamma(\alpha_0 + n)} \prod_{j=1}^{M} \frac{\Gamma(\alpha_j + t_j)}{\Gamma(\alpha_j)}, \quad (10)$$

where $t_j = \sum_{i=1}^{\nu} t_{i,j}$, $n = \sum_{j=1}^{M} t_j$. In the document clustering experiments we describe in Section 4, we set the $\alpha$'s in the prior of (10) to 1, which results in a uniform prior.

### 2.4.2 Model for a Flat Set of Clusters

Applying these multinomial and Dirichlet models to a set of clusters with the $(U, N)$ feature-set partition yields the following expression[14].

$$P(D|\Omega) = \left[ \frac{\Gamma(\gamma_U + \gamma_N)\Gamma(\gamma_U + t^U)\Gamma(\gamma_N + t^N)}{\Gamma(\gamma_U)\Gamma(\gamma_N)\Gamma(\gamma_U + t^U + \gamma_N + t^N)} \right]$$

$$\times \left[ \frac{\Gamma(\beta_0)}{\Gamma(\beta_0 + n^N)} \prod_{m=1}^{M_N} \frac{\Gamma(\beta_m + \tau_m)}{\Gamma(\beta_m)} \right] \quad (11)$$

$$\times \left[ \frac{\Gamma(\sigma)}{\Gamma(\sigma + \nu)} \prod_{k=1}^{K} \frac{\Gamma(\sigma_k + |D_k|)}{\Gamma(|D_k|)} \right]$$

$$\times \left[ \prod_{k=1}^{K} \frac{\Gamma(\alpha_0)}{\Gamma(\alpha_0 + n_k^U)} \prod_{j=1}^{M_U} \frac{\Gamma(\alpha_j + \tau_{k,j})}{\Gamma(\alpha_j)} \right],$$

where:

$t^U$ and $t^N$ are the total *useful* and *noise* term counts respectively for $D$.

$\tau_m$ is the total count for *noise* term $m$ in $D$.

$|D_k|$ is the total number of documents in cluster $k$.

$\tau_{k,j}$ is the total count of *useful* term $j$ in cluster $k$.

The $\gamma$'s, $\beta$'s, $\sigma$'s and $\alpha$'s are the hyperparameters associated with these various counts.

A complete derivation of (11) is provided in [14]. In writing (11) we have implicitly made the following assumptions: (1) The noise and useful features are conditionally independent, and (2) All parameter sets are independent.

### 2.4.3 Model for a Hierarchical Partition of a set of Documents

Equation (11) is the representation for a flat set of clusters. Our cluster-hierarchy model (fully expanded in the sense of (4) through (6)) will be significantly more complicated, containing a term like (10) for each noise distribution - $P(D_0^{N_0}|\Omega)$ and $\{P(D_k^{\tilde{N}_k}|\Omega)\}$ - and one such term for the *useful* feature distribution for each leaf cluster $\ell$: $P(D_\ell^{U_{P(\ell)}}|\Omega)$. As above we assume uniform priors for all parameter spaces.

## 3 Optimization Algorithm

The task for our optimization algorithm is that of finding the model structure $\hat{\Omega} \equiv \arg\max_\Omega P(D|\Omega)$ that maximizes equation (4), where the $\{P_k\}$ are fully recursively expanded and model forms appropriate to the application are substituted - e.g. multinomials in

our document-clustering application. This is an extremely difficult task even when we have closed-form expressions for the integrals, as we do in the case of multinomials. This is due to the enormous number of cluster hierarchies possible for $\nu$ objects. We use a reasonably efficient (albeit possibly suboptimal) approach consisting of the following two phase approach (explained in more detail below).

1. Perform a flat clustering where the number of clusters and the noise feature space is determined by optimizing equation (11).

2. Form a hierarchy from these clusters using a *modified hierarchical agglomerative clustering (MHAC) algorithm*.

### 3.1 Flat Clustering Algorithm

We perform flat clustering using the algorithm described in detail in [14, 13]. We review it briefly here. This algorithm finds the maximum-marginal-likelihood solution ($\hat{\Omega}$) of equation (1), using the models of Subsection 2.4, the EM algorithm and a *distributional-clustering* heuristic to reduce the feature-partition $(U, N)$ search space. The Bayesian *marginal-likelihood* provides a regularization criterion that allows the "natural" number of clusters to be determined automatically.

### 3.2 Modified Hierarchical Agglomerative Clustering Algorithm

Associated with the set of clusters obtained in this flat clustering scheme is a partition of the feature set into two subsets - $N$ and $U$. This partition is part of the optimal model structure $\hat{\Omega}$ for equation (1). The features in $N$ are global *noise* features in the sense of having a common distribution over all clusters. Our hierarchy model embodies a model structure $\Omega$ that also includes the possibility of features that have common distributions over proper sub-sets of clusters. To find the associated hierarchy and associated feature-set partitions we use a modified version of the hierarchical agglomerative clustering algorithm, which we denote by MHAC.

The HAC algorithm starts with singleton clusters (i.e., each data point is present in its own cluster) and is thus a computationally very expensive algorithm. As noted in [10], the lower parts of the dendrogram, created thus, provide no useful information. To overcome this problem Posse [10], generated an initial set of clusters using a minimum spanning tree which was then postprocessed using a series of heuristics. The resultant set of clusters is then used as input to the regular HAC algorithm. This is different than our approach where we generate the initial set of clusters using the EM algorithm and then merge them using the MHAC algorithm.

In MHAC we perform cluster merges that provide an *increase* in the *marginal likelihood of the hierarchy* (MLH) of equation (4). A merge of clusters is performed by first finding a feature sub-set that can be modeled as *noise* features for the pair of clusters under consideration. These *noise* features are modeled using a single set of parameters for these two clusters under consideration potentially increasing the MLH. The algorithm stops when none of the merges results in an increase in the MLH. The complete algorithm using both EM and the MHAC algorithm is as shown below.

> Repeat the following steps until only one cluster left or no increase in MLH
>> Identify two clusters, merging of which provides the largest increase in MLH
>>> For clusters $i$ and $j$ compute the increase in MLH by identifying features that can be made noise
>> Merge the clusters identified in the above step
>>> Compute sufficient statistics of the new cluster (sum of the sufficient statistics of the two clusters)

In the MHAC algorithm, we need to perform a search over the feature space to obtain a local set of noise features. To optimally accomplish the search over the feature space to obtain a local subset we would have to search over all possible $2^{N_{ij}}$ combinations of the available features. Since this is computationally intractable we resort to a greedy algorithm that evaluates one feature at a time and stops when the addition of a new feature does not provide an increase in marginal likelihood. To accomplish this we begin by first sorting the features in increasing order of $\Delta\hat{\xi}^j$ where $\hat{\xi}^j$ is the maximum likelihood estimate of the parameter for feature $j$ and $\Delta\hat{\xi}^j$ is the difference in $\hat{\xi}^j$ between the clusters being considered[2]. The algorithm keeps adding one feature at a time from this sorted list until the marginal likelihood starts decreasing.

It is worthwhile to note several important aspects about the MHAC algorithm. We begin with the initial value of MLH obtained as a result of the flat clustering algorithm[3]. We do not re-compute the complete MLH after every merge operation since all we are interested is the change in MLH.

---

[2]The rationale behind this heuristic is that features that have very similar ML estimates in both the clusters would be the best candidates for modeling as a single distribution

[3]Note that this is equivalent to a hierarchy with a root and as many children as clusters from the flat cluster

To compute this change in MLH we make use of the following theorem (proved in the Appendix), which states that this difference is dependent only on the sufficient statistics (i.e. cluster-wise term counts) of the two clusters and can be locally computed without accessing the actual data (document-wise term counts) in the clusters. This fact was also exploited in the maximum likelihood implementation by [9].

**Theorem 1** *For the models used in this work (multinomials with Dirichlet priors(M-D)), when merging two clusters, the change in LML is determined by only the projections of those clusters onto the subspaces spanned by the features that are chosen (as part of the merging process) to be common (i.e. noise) between those clusters. Thus the difference in LML between the two independent clusters (1 and 2) and their merged cluster (1,2) is given by:*

$$\log \left[ \frac{P_{(1,2)}\left(D_{(1,2)}^{N_{(1,2)}}|\Omega\right)}{P_1\left(D_1^{N_{(1,2)}}|\Omega\right) P_2\left(D_2^{N_{(1,2)}}|\Omega\right)} \right]. \quad (12)$$

Substituting our M-D models into (12) gives the expression for the difference in MLH:

$$\log \left[ \frac{\Gamma\left(\alpha_0 + t_1^{N_{(1,2)}}\right) \Gamma\left(\alpha_0 + t_2^{N_{(1,2)}}\right)}{\Gamma(\alpha_0 + t_{(1,2)}^{N_{(1,2)}})} \right] +$$
$$\sum_{j=1}^{|N_{(1,2)}|} \log \left[ \frac{\Gamma(\alpha_j + \tau_{(1,2),j})\Gamma(\alpha_j)}{\Gamma(\alpha_j + \tau_{1,j})\Gamma(\alpha_j + \tau_{2,j})} \right], \quad (13)$$

where we are using notation similar to that of (11) with the simple generalization that $t_k^V$ denotes the total count of feature-subset-$V$ terms in cluster $k$. Also the cardinality of the set $N_{(1,2)}$ is represented by $|N_{(1,2)}|$.

## 4 Experiments

In this section we describe the design of the experiments we use to evaluate our approach, using both synthetic and real-world data.

### 4.1 Data Sets

#### 4.1.1 Synthetic Data

We generated two synthetic data sets under the assumptions made in developing the model and objective function. To this end we defined two structures (shown in Figures 2 and 3) and for each structure we defined *noise* features at each intermediate node along with the associated parameters, as described in Section 2.

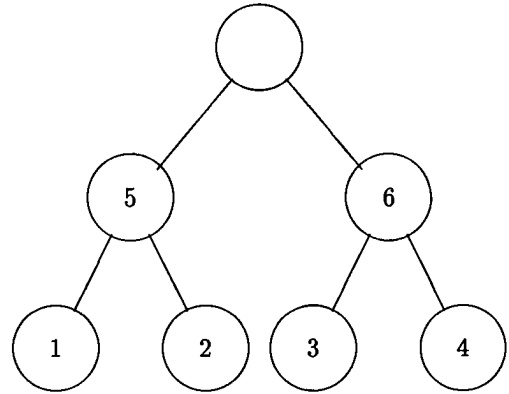

Figure 2: "Structure 1": A cluster hierarchy used in synthetic-data experiments.

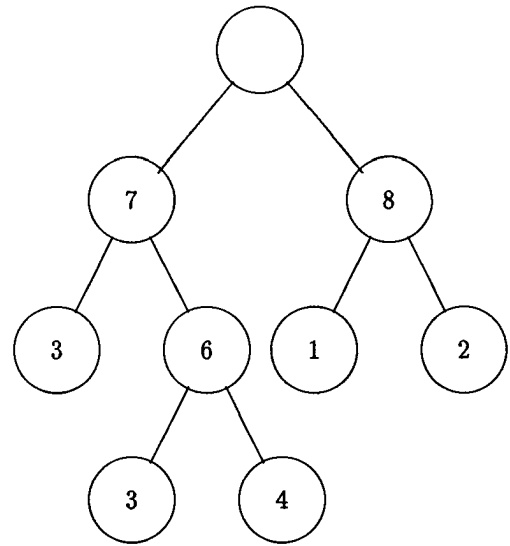

Figure 3: "Structure 2": A cluster hierarchy used in synthetic-data experiments.

In addition to generating data from a distribution that matched our model assumptions, we placed the following constraints to further limit the number of experiments.

- All data points (objects) are associated only with the terminal nodes[4] i.e. none are associated only with internal nodes.

- The total number of features is 50.

- The noise features for the intermediate levels (the $\{\tilde{N}_k\}$ of (2) and (3)) were generated un-

---

[4]Note that our objective function is capable of dealing with the situation where data exists in intermediate nodes - but we have ignored that aspect in this paper.

der the constraint that the total probability associated with useful features for each terminal node/cluster is at least 0.5.

Given these probabilities we sampled different numbers of data points (given in Tables 1 and 2) from this distribution for each of the structures shown in Figures 2 and 3.

### 4.1.2 Real-World Data - The TREC Document Collection

We have also applied our models and algorithms to a real world data set consisting of a collection of documents. An important practical problem today is the automatic creation of meaningful taxonomies from large collections of documents. The document collection we used was from the first 50 topics from the TREC collection. Each of these 50 topics, which form the leaf nodes in the taxonomy, is a member of one of 9 domains – and these domains form the next higher level of the taxonomy.

### 4.2 Description of Experiments

For all the data sets we used the MHAC algorithm to find our best estimate of $\hat{\Omega}$. The first step is the application of the EM algorithm[5]. For data from both structures we clustered the data sets into different numbers of clusters between 3 and 7 inclusive. For each of these numbers of clusters we ran the EM algorithm for 3 different initial seeds (these seeds were generated according to the method suggested in [9]). The optimum number of clusters is then selected based on the clustering that gave the highest marginal likelihood. These clusters are then used by the MHAC algorithm and the hierarchy is built starting at the bottom.

From these experiments we are interested in evaluating the algorithm's ability to do the following

1. Discover the right number of leaf clusters with the right assignments of data to clusters.

2. Uncover the hierarchical structure that was imposed.

3. Discover the appropriate noise features at each level of the hierarchical structure

4. Compute the Normalized mutual information

---

[5]Note that we did not generate any *noise* features at the root node (i.e. $N_0 = \emptyset$) thus precluding the search for noise features at this stage. For more information on such searches the reader is referred to work described in [14]

(NMI)[6] between the discovered hierarchy and the original (expert) hierarchy both at the leaves and at the intermediate level. NMI=1 corresponds to perfect agreement between class and cluster labels whereas NMI=0 corresponds to no correlation.

Evaluating the taxonomy generated by our approach for the real-world data is difficult. The TREC taxonomy that we chose has only two levels. The MHAC algorithm creates a hierarchy by merging the clusters two at a time and creating a dendrogram. To create an appropriate second level of 9 clusters, to correspond with the TREC taxonomy, we need to cut the dendrogram. To compute the following NMI measures we cut the MHAC generated dendrogram at the appropriate place. [7]

- NMI between the discovered hierarchy and the expert taxonomy at each of the two levels.

- NMI between the discovered hierarchy and the expert taxonomy at each of the two levels using a simple version of MHAC where we merge two clusters based on the least decrease in marginal likelihood

### 4.3 Experimental Results

#### 4.3.1 Results on Synthetic Data

The results of our experiments on synthetic data are shown in Tables 1 and 2. These tables present information about the number of features that were modeled as *noise* features at each level, the NMI at the leaves and the intermediate nodes and the number of merges the MHAC algorithm uncovered. For the EM algorithm we ran the synthetic data for different values of clusters ranging from 3 to 7 (for both the data sets) and for each number of clusters we ran the algorithm from 3 different starting points and chose the run with the highest marginal likelihood. For both the synthetic structures the EM algorithm found the right number of clusters. The NMI at the leaf is the normalized mutual information between the true assignment of data to clusters and the assignment of data to clusters as found by the EM algorithm. This value is referred to as NMI (leaf) in table 1. The NMI at the leaf is computed in the following manner. Consider the data-set 1. The 4 leaf clusters were used by the MHAC algorithm to perform merges and find the intermediate nodes. Noting that MHAC performs merges as long as the MHAC increases or no merges can be performed, we find in table 1 that with a sample size of 25000 the MHAC algorithm was able to perform only a single merge. All other sample sizes resulted in the maximum possible 2 merges. After the two merges we compute the NMI between the discovered intermediate clusters and the original assignment of data to the two intermediate clusters. This is referred to as NMI (inter) in table 1. A similar analysis, only with a maximum of 3 merges, is performed for data-set 2.

---

[6]A complete description of this approach is provided in [14].

[7]Note here that the MHAC algorithm performs merges

| size | MHAC | | NMI (leaf) | NMI (inter) |
|---|---|---|---|---|
| | Node 5 actual (found) | Node 6 actual (found) | | |
| 25000 | 15 (0) | 21 (7) | 0.7950 | NA |
| 50000 | 15 (4) | 21 (7) | 0.795 | 1.00 |
| 75000 | 15 (3) | 21 (9) | 0.7998 | 0.99 |
| 100000 | 15 (4) | 21 (9) | 0.7967 | 1.00 |

Table 1: Results for the synthetic data corresponding to Figure 2.

| size | MHAC | | | NMI (leaf) | NMI (int.) |
|---|---|---|---|---|---|
| | Node 6 act. (fnd) | Node 7 act. (fnd) | Node 8 act. (fnd) | | |
| 25000 | 24 (0) | 20 (0) | 15 (0) | 0.94 | 0.63 |
| 50000 | 24 (13) | 20 (8) | 15 (6) | 0.94 | 1.00 |
| 75000 | 24 (16) | 20 (9) | 15 (6) | 0.94 | 1.00 |
| 100000 | 24 (14) | 20 (9) | 15 (6) | 0.94 | 1.00 |

Table 2: Results for the synthetic data corresponding to Figure 3.

### 4.3.2 Results on a Real Document Collection

For the real-world data set we first performed a feature extraction step where the features were extracted as described in [4, 11]. The feature selection was performed using a distributional clustering algorithm as described in [14]. After feature selection we were left with a total of 14772 tokens. To enable appropriate comparison against the TREC taxonomy we clustered the 16801 documents into 50 clusters for 3 different starting points (initial partitions) and chose the run that gave us the highest marginal likelihood). These 50 clusters were then merged using two versions of the MHAC algorithm, one with feature selection (MHAC-FS) and one without feature selection (MHAC-noFS). The version without feature selection simply merged clusters based on the least decrease in marginal likelihood. The results of these experiments (see Table 3) are discussed in the following subsection.

| Level | NMI MHAC-FS | NMI MHAC-noFS |
|---|---|---|
| Level 2 | 0.513 | 0.513 |
| Level 1 | 0.300 | 0.280 |

Table 3:

### 4.4 Discussion

As seen from Tables 1 and 2 the EM+MHAC algorithm was able to find the structure more easily in the first data set as opposed to the second. Importantly, EM found the right number of clusters in both the structures and for all sets of samples. This can be partly explained by the fact that the probability associated with useful features at the leaf nodes was forced to be greater than 0.5. If the number of useful features at the leaf level is very small, it can be difficult for EM to recover the correct number of leaf nodes and we may have to resort to other approximations. We have not pursued this set of experiments here however. The NMI is greater at the intermediate level indicating that the confusion in assignments of data at the leaf nodes, by EM, was possibly caused by the noise features at the intermediate level. Another important factor is that the NMI at the second level for structure 2 is perfect for all sample sizes greater than 25000. Unlike the case of the first structure, where the MHAC-FS algorithm only found a single merge (all other merges did not result in an increase in the MLT), in the structure 2 for 25000 samples the algorithm performed wrong merges. This is reflected in the drastic drop in the value of NMI for this data set (shown in Table 2).

With the real-world data set we note that the MHAC-FS provides a slight improvement over MHAC-noFS. An added advantage of the MHAC-FS algorithm is that the model also provides us with a list of common features at each intermediate node which, for the application of taxonomies, can be used as labels to describe the intermediate nodes.

We also conducted an informal user study where users were given 20 randomly chosen documents from the original collection of 16801 documents. One set of users were required to classify these documents into the hierarchy generated by EM+MHAC (completely automaticly) while another set were asked to classify the same documents into the expert hierarchy (the TREC hierarchy). Users were evaluated both on the number of misclassifications and on the amount of time taken to classify the documents. Users on an average took 15 minutes more to classify documents into the hierarchy generated using EM+MHAC with a 7.5% lower accuracy in classification. The total time taken to classify 20 documents into the TREC taxonomy was 45 minutes.

## 5 Summary and Future Work

In this paper we have introduced a new model for arranging objects in a natural hierarchy. The key idea that is exploited is the fact that certain features may have a common distribution across some clusters while having different distribution in others. This model has several applications including the important problem of automatic taxonomy generation. The objective function associated with our model is based on a rigorous Bayesian analysis and we have derived a closed-form for the Bayesian marginal likelihood of the hierarchy under certain assumptions. We then provide a novel algorithm to find an approximate solution to this objective function. This algorithm works in two stages - in the first stage a flat set of partitions is generated and then a modified version of the HAC algorithm is used to model the intermediate nodes and the features that are common across its children. We have then applied this model to two synthetic datasets and to one real world data-set.

Despite our encouraging results, several research possibilities remain open both in the model and the algorithms. Extensions to the hierarchical model to include more complicated structures such as *directed acyclic graphs* (DAGs) need to be investigated. Further, all of our experiments use a uniform prior for $\pi(\xi)$ - other priors need to be investigated. In the algorithm several issues need to be addressed. For example there is the need to overcome the MHAC limitation of merging two clusters - i.e., noise features are modeled across multiple children ($> 2$). We plan to investigate these extensions in our future work.

Acknowledgement: The authors would like to thank Yael Ravin and Roy Byrd for free use of Textract and all the timely help provided.

## A  Proof of Theorem 1: Analysis of Merging Process

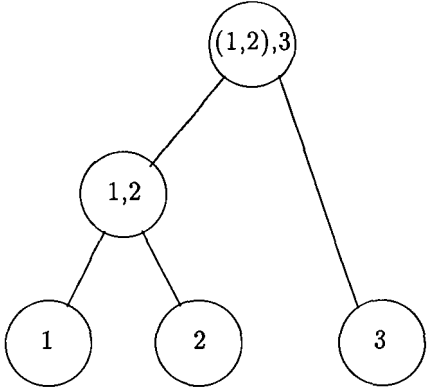

Figure 4: Diagram of the simple cluster hierarchy we use to analyze the merging process.

To prove this theorem we analyze the merging process associated with the simple hierarchy shown in Figure 4.

The initial feature-set partition associated with the initial flat clustering (consisting of the three independent clusters 1,2 and 3) is given by:

$$T \to (N, U)$$

For simplicity we assume that $N = \emptyset$ because these features have no effect on the merging process. This is due to the fact that our merging process is greedy and the decision to make these features global *noise* features is not revisited.

In what follows we adopt the following simplifying notation. For cluster $k$ and feature subspace $V$:

$$< k|V > \equiv P(D_k^V | \Omega).$$

Note that because of the specific model that we use (M-D) the following identity holds for any two non-overlapping subsets $V_a$ and $V_b$ composing $V$.

$$< k|V > = < k|V_a >< k|V_b > . \qquad (14)$$

The following proof holds not just for our M-D models, but for any models satisfying (14).

### A.1  First Merge $1 + 2 \to (1, 2)$

Initially (before merging) the marginal likelihood of the two clusters is given by $< 1|U >< 2|U >$ and after the merge it is given by

$$< (1,2)|N_{(1,2)} >< 1|U_{(1,2)} >< 2|U_{(1,2)} > . \qquad (15)$$

The difference in LML is simply the log of the ratio of these two quantities. To simplify this ratio we further decompose $< 1|U >$ and $< 2|U >$. Remember that

$$U = N_{(1,2)} \cup U_{(1,2)}.$$

This fact combined with the decomposition rule (14), gives us:

$$< 1|U > = < 1|N_{(1,2)} >< 1|U_{(1,2)} >$$

and a similar expression for $< 2|U >$. Substituting these expressions into likelihood ratio associated with the merge and cancelling terms yields:

$$\frac{< (1,2)|N_{(1,2)} >}{< 1|N_{(1,2)} >< 2|N_{(1,2)} >}. \qquad (16)$$

This proves the theorem, but we go on to consider the second merge $(1, 2) + 3 \to ((1, 2), 3)$ because there is a certain restriction associated with it.

### A.2  Second Merge: $(1, 2) + 3 \to ((1, 2), 3)$

Simply applying the analysis that led to (16) to our second merge yields:

$$\frac{< ((1,2),3)|N_{((1,2),3)} >}{< (1,2)|N_{((1,2),3)} >< 3|N_{((1,2),3)} >}. \qquad (17)$$

It should be pointed out that part of the merging process is determining the associated feature partition. Thus various possibilities of $N_{((1,2),3)}$ are tried and the one giving the highest ML for the merged cluster is chosen. Any possible partition of $U$ can be chosen when merging the original clusters from the flat clustering. Whenever higher-level clusters (e.g. (1,2)) are included in the merge, however, not all such feature partitions are allowed.

Because we perform a forward-only greedy merging process, once the feature set $U_{(1,2)}$ has been selected as part of the $1 + 2 \to (1, 2)$ merge, it will be fixed from that point on throughout the merging process. The implication of this is that the features comprising $U_{(1,2)}$ are ineligible for inclusion in $N_{((1,2),3)}$. In general for a merge being considered at any point in the hierarchy all features that have been included in any *useful* feature set as part of previous (lower-level) merges are ineligible to be included in the noise feature set associated with the higher-level merge.